# Complexity Analysis and Variational Inference for Interpretation-based Probabilistic Description Logics

**Fabio Gagliardi Cozman** and **Rodrigo Bellizia Polastro**
Escola Politécnica, Universidade de São Paulo - São Paulo, SP - Brazil

## Abstract

This paper presents complexity analysis and variational methods for inference in probabilistic description logics featuring Boolean operators, quantification, qualified number restrictions, nominals, inverse roles and role hierarchies. Inference is shown to be **PEXP**-complete, and variational methods are designed so as to exploit logical inference whenever possible.

## 1 Introduction

In this paper we investigate *probabilistic description logics* that are based on the well-known description logic $\mathcal{ALC}$. In $\mathcal{ALC}$ one deals with individuals, concepts, roles, Boolean operators, and restricted forms of quantification [1]. For example, the concept Fireman denotes the set of firemen, and Vegetarian $\sqcap$ $\forall$buyFrom.Fireman denotes a set of individuals who are vegetarian and buy only from firemen. We can also have assertions such as Fireman(John), stating that John belongs to Fireman.

We start from *credal* $\mathcal{ALC}$, a probabilistic description logic we have introduced previously [8, 33]. Credal $\mathcal{ALC}$, referred to as CR$\mathcal{ALC}$, mixes constructs of $\mathcal{ALC}$ with features of relational Bayesian networks. Indeed most of this paper can be read as the study of those relational Bayesian networks that can be expressed with (variants) of $\mathcal{ALC}$.

In CR$\mathcal{ALC}$ one can have probabilistic assessments such as $P(C|D) = \alpha$ for concepts $C$ and $D$, or $P(r) = \beta$ for role $r$. The semantics of these assessments is roughly given by:

$$\forall x : P(C(x)|D(x)) = \alpha, \quad \forall x, y : P(r(x,y)) = \beta.$$

Credal $\mathcal{ALC}$ is attractive because its semantics allows reasonably flexible probabilistic assessments *and* the calculation of probabilities over assertions; the language stays both close to the power of $\mathcal{ALC}$ and to the clarity of relational Bayesian networks. Section 2 reviews work on probabilistic description logics and in particular CR$\mathcal{ALC}$.

To illustrate the main features of CR$\mathcal{ALC}$, consider an example whose syntax should be reasonably easy to grasp. Take the Kangaroo ontology, a small collection of facts about animals.[1] Consider a probabilistic version as follows [33]:
$P(\text{Animal}) = 0.9$, $P(\text{Rational}) = 0.6$, $P(\text{hasChild}) = 0.3$,
Human $\equiv$ Animal $\sqcap$ Rational,
Beast $\equiv$ Animal $\sqcap$ ¬Rational,
Parent $\equiv$ Human $\sqcap$ $\exists$hasChild.Human,
$P(\text{Kangaroo}|\text{Beast}) = 0.4$, $P(\text{Kangaroo}|\neg\text{Beast}) = 0.0$,
MaternityKangaroo $\equiv$ Kangaroo $\sqcap$ $\exists$hasChild.Kangaroo.

The most basic problem in CR$\mathcal{ALC}$ is to compute the probability of an assertion, possibly conditional on other assertions. For instance, we might be interested in computing $P(\text{MaternityKangaroo}(\text{Tweety}))$ given the probabilistic Kangaroo ontology. In Section 3 we present complexity analysis for this *inference* problem. Variants of CR$\mathcal{ALC}$ are also analyzed in Section 3, both by removing some constructs from the language, and by adding constructs such as numeric restrictions, inverse roles, and nominals.

In Section 4 we propose a variational scheme for inference in CR$\mathcal{ALC}$ and its variants, designed to exploit Boolean operators (through logical inference) in a set of sentences. We often adopt a *Bayesian assumption* that guarantees uniqueness of probability values, but we also briefly examine failures of this assumption. Experiments described in Section 4 show the value of the algorithms in practice.

## 2 Probabilistic description logics

The combination of probability and logic has a long history, with much recent activity [16, 17, 34]. In particular, there has been significant interest in probabilistic description logics [29]. The next two paragraphs summarize recent activity in this topic; a more detailed review and comparison can be found in our previous publication [8, Sec.2].

We can divide probabilistic description logics into logics that assign probabilities to subsets of the domain [13, 14, 18, 21, 25, 28] and to subsets of interpretations [7, 10], with some logics in between [37]. In a domain-based semantics an assessment such as $P(\mathsf{Fireman}) = 1/2$ means that the probability mass over the set of all firemen is half. The challenge then is to define probabilities for an assertion: the probability of the set of individuals who are firemen does not constrain the probability that a particular individual is a fireman. This is indeed the old philosophical problem of *direct inference* [26]. Hence logics with domain-based semantics either do not allow probabilities of assertions to be expressed, or resort to non-standard forms of entailment [21, 28]. This is the reason why we developed CR$\mathcal{ALC}$ as an interpretation-based probabilistic logic.

---

[1] Distributed with the CEL System for logical reasoning, at http://lat.inf.tu-dresden.de/~meng/ontologies/kangaroo.cl.

We can alternatively divide probabilistic description logics into logics that allow independence relations to be organized into graphs [7, 13, 25], and logics that do not resort to independence relations [14, 18, 21, 28]. Logics in the first group usually assume some Markov condition, and assume that probabilities are *uniquely* defined for any valid sentence. Logics in the second group allow probability intervals (sometimes sets of probabilities) to be associated with sentences. We think that independence relations are powerful constraints that should be used whenever possible, and for this reason CR$\mathcal{ALC}$ has many similarities to logics in the first group, even though it does not mandate uniqueness for probabilities. In particular, CR$\mathcal{ALC}$ shares many features with PR-OWL [7] as both have interpretation-based semantics and use graph-theoretical tools.

We now turn to a more precise description of CR$\mathcal{ALC}$, starting with background on $\mathcal{ALC}$ [36]. Throughout, $a, b$ are individuals; $A, B, C, D$ are concepts; $r, s$ are roles. A *terminology* contains *inclusions* $C \sqsubseteq D$ and *definitions* $C \equiv D$. A concept can be a concept name or, recursively, a *conjunction* ($C \sqcap D$), disjunction ($C \sqcup D$), negation ($\neg C$), existential restriction ($\exists r.C$), value restriction ($\forall r.C$). Concept $C$ *directly uses* $D$ if they appear respectively in the left and right hand sides of an inclusion/definition. The relation *uses* is always the transitive closure of *directly uses*. A terminology is *acyclic* if no concept *uses* itself in an inclusion/definition. An *Abox* contains assertions such as $C(a)$, $r(a, b)$. An assertion $C(a)$ *directly uses* assertions of concepts and roles directly used by $C$ instantiated respectively by elements and pairs of elements from $\mathcal{D}$. The semantics of $\mathcal{ALC}$ is given by a set $\mathcal{D}$, the *domain*, and a mapping $\mathbb{I}$, the *interpretation*. This mapping takes each individual to an element of the domain, each concept name to a subset of the domain, each role name to a binary relation on $\mathcal{D} \times \mathcal{D}$. An interpretation is extended to other concepts: $\mathbb{I}(C \sqcap D) = \mathbb{I}(C) \cap \mathbb{I}(D)$, $\mathbb{I}(C \sqcup D) = \mathbb{I}(C) \cup \mathbb{I}(D)$, $\mathbb{I}(\neg C) = \mathcal{D} \backslash \mathbb{I}(C)$, $\mathbb{I}(\exists r.C) = \{x \in \mathcal{D} | \exists y : (x, y) \in \mathbb{I}(r) \wedge y \in \mathbb{I}(C)\}$, $\mathbb{I}(\forall r.C) = \{x \in \mathcal{D} | \forall y : (x, y) \in \mathbb{I}(r) \rightarrow y \in \mathbb{I}(C)\}$. We have $C \sqsubseteq D$ if and only if $\mathbb{I}(C) \subseteq \mathbb{I}(D)$, and $C \equiv D$ if and only if $\mathbb{I}(C) = \mathbb{I}(D)$. There are translations of these constructs into modal and first-order logic [1]; we often treat a concept $C$ as a unary predicate $C(x)$, a role $r$ as a binary predicate $r(x, y)$, and restrictions as quantifiers.

In CR$\mathcal{ALC}$, we have all constructs of $\mathcal{ALC}$ but we only allow concept names in the left hand side of inclusions/definitions. Additionally, we allow three kinds of probabilistic assessments, where $C$ is a concept name, $D$ is a concept, $r$ is a role name:

$$P(C) \in [\underline{\alpha}, \overline{\alpha}], \quad (1)$$
$$P(C|D) \in [\underline{\alpha}, \overline{\alpha}], \quad (2)$$
$$P(r) \in [\underline{\beta}, \overline{\beta}]. \quad (3)$$

In Expression (2), $C$ *directly uses* conditioning concepts. We write $P(C|D) = \underline{\alpha}$ when $\underline{\alpha} = \overline{\alpha}$, $P(C|D) \geq \underline{\alpha}$ when $\underline{\alpha} < \overline{\alpha} = 1$, and so on. No concept is allowed to use itself, neither through deterministic nor through probabilistic inclusions/definitions; this guarantees acyclicity.

The semantics is based on probabilities over interpretations; that is, we take that measures are assigned to the set of all interpretations. The semantics of Expression (1) is: for any $x \in \mathcal{D}$, the probability that $x$ belongs to the interpretation of $C$ is in $[\underline{\alpha}, \overline{\alpha}]$. That is,

$$\forall x \in \mathcal{D} : P\Big(\big\{\mathbb{I} : x \in \mathbb{I}(C)\big\}\Big) \in [\underline{\alpha}, \overline{\alpha}].$$

An informal and intuitive way to express the semantics is

$$\forall x \in \mathcal{D} : P(C(x)) \in [\underline{\alpha}, \overline{\alpha}].$$

The semantics of Expressions (2) and (3) is then:

$$\forall\, x \in \mathcal{D} \;:\; P(C(x)|D(x)) \in [\underline{\alpha}, \overline{\alpha}],$$
$$\forall\, (x, y) \in \mathcal{D} \times \mathcal{D} \;:\; P(r(x, y)) \in [\underline{\beta}, \overline{\beta}].$$

As usual in interpretation-based probabilistic logics [3], CR$\mathcal{ALC}$ requires that all individuals be *rigid* (an individual corresponds to the same element of the domain across interpretations). Similarly to other probabilistic description logics [7, 13, 25], CR$\mathcal{ALC}$ adopts a (two-part) Markov condition that is best formulated using indicator functions.[2] First, for every concept $C$ and for every $x \in \mathcal{D}$, the indicator function of $C(x)$ is independent of the indicator function of every assertion of a concept that does not use $C(x)$, given the indicator function of assertions of concepts that $C$ directly uses. Second, for every role $r$ and for every $(x, y) \in \mathcal{D} \times \mathcal{D}$, the indicator function of $r(x, y)$ is independent of every indicator function of assertions, except those assertions of concepts that use $r(x, y)$. This closes the specification of CR$\mathcal{ALC}$.

We define the *t-network* of a terminology to be a directed acyclic graph where each node is a concept name or a restriction or a role name. If concept $C$ directly uses other concept/role names, these names appear as *parents* of $C$ in the t-network. To simplify the presentation, we assume that only a concept name $C$ can appear in a restriction $\forall r.C$ or $\exists r.C$ (without loss of generality, as $C$ can be defined through other concepts). A restriction $\forall r.C$ or $\exists r.C$ has parents $r$ and $C$ in the t-network. Figure 1 shows the t-networks for the Kangaroo ontology (Section 1) and for the following simple example.

---

[2]The indicator function of a grounded relation $C(x)$ or $r(x, y)$ yields 1 if the grounded relation holds and 0 otherwise.

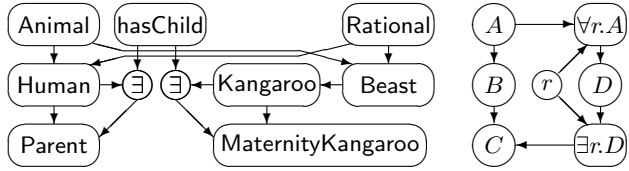

Figure 1: Left: t-network for Kangaroo ontology (Section 1). Right: t-network for terminology $\mathbb{T}_u$ (Example 1).

**Example 1 (from [8])** Consider terminology $\mathbb{T}_u$:
$P(A) = 0.9$, $\quad B \sqsubseteq A$, $\quad P(B|A) = 0.45$, $\quad D \equiv \forall r.A$,
$C \equiv B \sqcup \exists r.D$, $\quad P(r) = 0.3$.

Consider a terminology $\mathbb{T}$, a set $\mathcal{C}$ containing the concept names and the restrictions in $\mathbb{T}$, a set $\mathcal{R}$ containing the role names in $\mathbb{T}$, and a finite domain $\mathcal{D}$. Denote by $\mathrm{pa}(C(a))$ the set of all indicator functions of assertions that $C(a)$ directly uses ($C$ may be a restriction). Under the Markov condition, any joint probability distribution for the set $\mathbf{X}$ of indicator functions of assertions from $\mathbb{T}$ and $\mathcal{D}$ has the form

$$P(\mathbf{X}) = \prod_{C \in \mathcal{C}; x \in \mathcal{D}} P(C(x)|\mathrm{pa}(C(x))) \times \prod_{r \in \mathcal{R};(x,y) \in \mathcal{D} \times \mathcal{D}} P(r(x,y)). \quad (4)$$

Throughout the paper we do not differentiate a grounded relation from its indicator function: $C(x)$ and $r(x,y)$ stand for indicator functions of binary and unary relations in Expression (4). We assume throughout the *unique-name* assumption: if $a$ and $b$ are distinct names for individuals, their interpretations are distinct. We also contemplate two other assumptions. The *confined-domain* assumption says that $\mathcal{D}$ has finite known cardinality, denoted by $N$. The *uniqueness* assumption says that there is a precise assessment for each role name ($P(r) = \beta$), and for each concept name we have either a definition $C \equiv D$, or a single assessment $P(C) = \alpha$, or a pair of assessments $\{P(C|C') = \alpha', P(C|\neg C') = \alpha''\}$ where $C'$ must be a concept name (that can be defined elsewhere). Under these three assumptions, collectively referred to as the *Bayesian assumption*, every CR$\mathcal{ALC}$ terminology defines a unique *relational Bayesian network* [22] whose grounding is a Bayesian network given by Expression (4). Logics such as P-CLASSIC and PR-OWL adopt similar assumptions.

An *inference* in CR$\mathcal{ALC}$ is the computation of $P(A_0(a_0)|\mathcal{A})$ given a terminology and an Abox $\mathcal{A} = \{A_j(a_j)\}_{j=1}^M$. For instance, given the terminology $\mathbb{T}_u$ in Example 1 we have the trivial inferences $P(A(a)) = 0.9$ and $P(A(a)|A(a)) = 1$ for any $N$.

## 3 Expressivity, complexity, and CR$\mathcal{HO}_r\mathcal{IQ}$

We now examine the interplay between expressivity and complexity in CR$\mathcal{ALC}$ and its extensions. Besides usual classes **P**, **NP**, **PSPACE** and **NEXP**, we use three other classes. A language $L$ is in class **PP** if there is a nondeterministic Turing machine $\mathbb{M}$ such that $x \in L$ if and only if more than half of the computations of $\mathbb{M}$ on input $x$ end up accepting, when $\mathbb{M}$ has a *polynomial-time* bound [32]; $L$ is in **PPSPACE** if the same definition is used but we replace *polynomial-time* by *polynomial-space* [31]; and $L$ is in **PEXP** if the same definition is used but we replace *polynomial-time* by *exponential-time* [5].

Define $Q \doteq P(A_0(a_0)|\mathcal{A})$ for an Abox $\mathcal{A}$, and denote by $\mathsf{INF}_B(Q)$ the decision problem that returns YES if $Q > 1/2$ and NO otherwise, under the Bayesian assumption. The assumption that numeric parameters are coded in unary is common in description logic research [1]. Even though in the present context unary specification seems rather artificial because probabilities are normally coded in binary, the following easy proposition is worth stating.

**Proposition 1** *If $N$ is given in unary and probabilities in binary, $\mathsf{INF}_B(Q)$ is a **PP**-complete problem in CR$\mathcal{ALC}$.*

*Proof.* Membership: Propositionalize terminology into Bayesian network with $N|\mathcal{C}| + N^2|\mathcal{R}|$ variables, and run probabilistic inference. Hardness: for $N = 1$ the inference is **PP**-complete (Bayesian network inference) [27]. □

A much more interesting question is the complexity of inferences for $N$ in *binary*, where the domain is exponentially larger than its description. One might conjecture that, as satisfiability in $\mathcal{ALC}$ is in **PSPACE**, inference in CR$\mathcal{ALC}$ should be in **PPSPACE**. One the other hand, as Jaeger's important previous analysis [24] indicates that (unless **ETIME**=**NETIME**) there must be model representation systems for which inference is not in **P** with respect to $N$ in unary, one might suspect that inference with $N$ in binary should take us to exponential time complexity of some sort. The next theorem offers the precise completeness result for inference; the proof, summarized in the Appendix, is somewhat long and contains ideas that may be of general interest.

**Theorem 1** *If all numbers are given in binary, $\mathsf{INF}_B(Q)$ is a **PEXP**-complete problem in CR$\mathcal{ALC}$.*

We thus have a clear difference between probabilistic reasoning with an enumerated domain (Proposition 1) and with a compactly specified domain (Theorem 1). One might try to reduce the complexity of $\mathsf{INF}_B(Q)$ by starting with a description logic simpler than $\mathcal{ALC}$, for instance by discarding some operators and negation [2]. However, one can "probabilistically negate" a concept $C$ by creating a new concept $C'$ that has probability 0 when $C$ obtains and probability 1 otherwise. Thus the proof of Theorem 1 can be reconstructed if we restrict CR$\mathcal{ALC}$ even to the logical constructs of the simple logic $\mathcal{EL}$ (conjunction and existential quantification) [2]. Hence $\mathcal{ALC}$ seems to be the minimal start for a probabilistic description logic, and the gap between **PP** and **PEXP** does seem to be the minimal gap between an essentially propositional and a truly relational probabilistic representation system.

We now consider the opposite direction; that is, the complexity of inference in logics that extend CR$\mathcal{ALC}$. Consider the following constructs. A *qualified number restriction* $(\geq k \ r.C)$ has semantics $\mathbb{I}(\geq k \ r.C) = \{x \in \mathcal{D} : \#\{y \in \mathcal{D} : (x,y) \in \mathbb{I}(r) \wedge y \in \mathbb{I}(C)\} \geq k\}$, where $\#(\cdot)$ yields the cardinality of the set. Number restrictions $(\leq k \ r.C)$ and $(= k \ r.C)$ are defined similarly. An *inverse role* $r^-$ is interpreted by replacing $r^-$ and $(x,y)$ respectively by $r$ and $(y,x)$ in the semantics. A *role hierarchy* is based on inclusions $r \sqsubseteq s$, and a probabilistic version consists of assessments $P(r|s) \in [\underline{\beta}', \overline{\beta}']$ and $P(r|\neg s) \in [\underline{\beta}'', \overline{\beta}'']$. Such probabilistic role hierarchies demand some strenghtening of the uniqueness condition: we must assume that every role $r$ must be either associated with an assessment $P(r) = \beta$ or a pair $\{P(r|s) = \beta', P(r|\neg s) = \beta''\}$.

It is also possible to add *nominals* to CR$\mathcal{ALC}$; a nominal is an individual name identified with a concept. Nominals would add enormous expressivity to CR$\mathcal{ALC}$: for instance, one might express probabilities for a particular individual $a$ through the assessment $P(C|\{a\}) = \alpha$. However it does not seem reasonable to ask for assessments such as $P(\{a\}) = \alpha$, meaning $\forall x \in \mathcal{D} : P(\{a\}(x)) = \alpha$, because this sentence is clearly false for every $x$ if $\alpha \in (0,1)$. Either the syntax regarding nominals and probabilities must be significantly changed, so that nominals cannot be assigned probabilities, or the inference algorithms presented later would have to be changed so as to detect inconsistencies between assessments and nominals. In this paper we do not attempt to model nominals in their full generality; instead we only allow nominals in restrictions such as $\forall r.\{a\}$, $\exists r.\{a\}$ and $\geq kr.\{a\}$. Such constructs seem to capture significant portion of the practical use of nominals.

As usual in description logics, we indicate numeric restrictions by the letter $\mathcal{Q}$; inverse roles by $\mathcal{I}$; role hierarchies by $\mathcal{H}$. We use $\mathcal{O}_r$ to indicate the restricted use of nominals described in the previous paragraph. Whenever possible we remove the letters $\mathcal{ALC}$ from names, so CR$\mathcal{ALC}$ with the additional features just mentioned is referred to as CR$\mathcal{HO}_r\mathcal{IQ}$. This logic contains most of $\mathcal{SHOIQ}$, a logical basis for the OWL language [19].[3] We have:

**Theorem 2** *If all numbers are given in binary*, $\mathsf{INF}_B(Q)$ *is a* **PEXP**-*complete problem in logics whose features contain* CR$\mathcal{EL}$ *and are contained in* CR$\mathcal{HO}_r\mathcal{IQ}$.

---

[3] To reach $\mathcal{SHOIQ}$ we would need *transitive roles*; this seems to require new ideas as transitivity violates the Markov condition (groundings of the same role are dependent given transitivity).

*Proof.* Membership: argument in Theorem 1 is not affected by the new features. Hardness follows from Theorem 1. □

To conclude this section, we briefly comment on failures of the confined-domain and uniqueness assumptions. By retaining the confined-domain assumption and dropping the uniqueness assumption, the decision problem "$\min Q > 1/2$?" belongs to **NEXP$^{\text{PEXP}}$**, as there are exponentially many choices of probabilities (only the vertices of the sets of distributions can generate minimizing/maximizing probabilities [15], and there are exponentially many vertices) and for each one of these choices, an oracle in **PEXP** yields the answer. One might adopt an *homogeneity* assumption [8] prescribing that the selection of a probability (within a probability interval) should be constant across individuals; this assumption moves the decision problem to **NP$^{\text{PEXP}}$**, as there are polynomially many choices concerning probabilities, followed by an oracle in **PEXP**. Other situations are more difficult to analyze. For instance, suppose we adopt uniqueness and take $N$ to be countably infinite. Using results by Jaeger [23], we know that CR$\mathcal{ALC}$ has a 0/1-law such that the probability of every restriction goes to 0 or 1 as $N$ grows without bound [8]; if we could determine the limiting value of probabilities for quantifiers, then the grounded network would decompose into a set of independent Bayesian networks and inference would be **PP**-complete. Of course, determining the limiting values for restrictions is not an easy matter [23], so our point is just to describe it as a difficult open problem. Likewise, we do not have complexity results for the challenging problems that arise when $N$ is unconstrained. In Section 4.3 we produce approximate methods for such inferences. Indeed, we focus on approximate inference in the remainder of the paper, as it should be clear that applications will often have to resort to approximations.

## 4 Variational inference for CR$\mathcal{HO}_r\mathcal{IQ}$

We now look into methods that approximate $Q = P(A_0(a_0)|\mathcal{A})$, where the assertions in $\mathcal{A}$ and the nominals in the terminology refer to individuals $\{a_1, \ldots, a_M\}$. Define $\mathcal{D}' \doteq \{a_1, \ldots, a_M\}$ and assume, without loss of generality, that $a_0 \in \mathcal{D}'$. If we ground a terminology, we obtain a directed acyclic graph with $N$ slices: The *slice of a* is the set of indicator functions for grounded concepts $C(a)$ and grounded roles $r(a,x)$ as $x$ ranges over $\mathcal{D}$. Only $M$ slices refer to named individuals in $\mathcal{D}'$; the other $N - M$ slices are identical for all purposes, and we can lump them into a single parameterized slice. The network with a slice per named individual $a_i$ and an additional slice for a "generic" element $x \in \mathcal{D}\setminus\mathcal{D}'$ is called the *shattered network* [8].[4] Figure 2 (left) shows the shattered network for terminol-ogy $\mathbb{T}_u$ in Example 1 and query $P(A(a_0)|C(a_0))$; there is a slice for $a_0$ (the only named individual in the query) and a single slice for $x \in \mathcal{D}\backslash\mathcal{D}'$. An additional variable $y \in \mathcal{D}\backslash\mathcal{D}'$ is used to represent groundings of roles in the parameterized slice.

---

[4] This network is a representation for the *shattering* operation in first-order variable elimination [11]; the relationship between shattering and shattered networks is given by [8, Thm. 1].

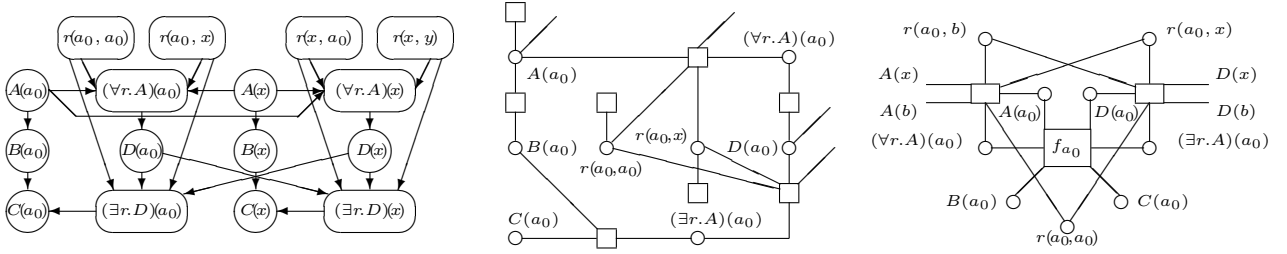

Figure 2: Left: shattered network for query $P(A(a_0)|C(a_0))$, terminology $\mathbb{T}_u$ from Example 1. Middle: fragment of the factor graph corresponding to the eight leftmost nodes of the shattered network. Right: factor graph for the slice of $a_0$.

Parts of the shattered network can be eliminated using d-separation, possibly leading to substantial savings. For instance, if d-separation eliminates all quantifiers and interval-valued assessments, we face a Bayesian network inference. We assume that such easy cases are detected before our variational method is employed.

In what follows we focus on universal/existential restrictions for the sake of space. Numeric restrictions and role hierarchies can be handled with straightforward modifications, while inverse roles can also be handled by the same methods but the equations presented later must be changed significantly. Finally, each nominal $a$ that appears in restrictions can be dealt with by creating a concept $\{a\}$ such that $\{a\}(a)$ holds while $\{a\}(x)$ does not hold for $x \neq a$; the shattered network now has all nodes $\{a\}(x)$ clamped to false in its parameterized portion, and all techniques presented later apply. (A different scheme for inference in CR$\mathcal{ALC}$ with restricted nominals can be found in [33].)

### 4.1 Inference under the Bayesian assumption

Under the Bayesian assumption every grounded network is a Bayesian network. Such networks tend to be densely connected due to the presence of restrictions. Indeed, even the state-of-art inference engine ACE [6] fails to handle the grounded network of Example 1 for $N = 10$. We have previously suggested the use of Loopy Belief Propagation (LBP) for approximate inference in large grounded networks [8, 33]. The key observation is that many messages in LBP are repeated across the grounded network, and can therefore be lumped into parameterized messages. The resulting parameterized LBP is similar to the *lifted* propagation scheme of Singla and Domingos [38]. The difference is that Singla and Domingos' algorithm applies to a general logic and offers few guarantees regarding the number of lifted messages to be exchanged in each iteration, while the number of parameterized messages is fixed in our previous method, due to the regularity of CR$\mathcal{ALC}$ terminologies. To understand the idea behind parameterized LBP, suppose that in Example 1 we are interested in $P(D(a_0)|B(a_0))$. In LBP we need only to send messages from groundings of $A$ and $r$ to $(\forall r.A)(a_0)$, and from this latter node to $D(a_0)$. Node $A(a_0)$ sends a *single* message to $(\forall r.A)(a_0)$, while node $A(x)$ sends $N - 1$ *identical* messages to $(\forall r.A)(a_0)$. Then $P((\forall r.A)(a_0))$ is

$$(1-P(r)(1-P(A)))^{N-1}(1-P(r)(1-P(A(a_0)|B(a_0)))).$$

This parameterized LBP scheme, where parameterized messages are raised to appropriate powers because they stand for sets of messages, is iterated until convergence.

In this section we present improvements to this parameterized LBP scheme, motivated by two observations. First, while LBP usually displays excellent empirical performance, in LBP all nodes are treated alike, thus forsaking the possible exploitation of logical reasoning in a t-network. We expect that applications for probabilistic description logics will contain quite a few deterministic definitions and Boolean operators. Hence it is important to develop probabilistic inference that can exploit logical inference whenever appropriate. Second, we note that most terminologies in the literature are represented as relatively sparse graphs; we expect that applications will contain t-networks such that the case $N = 1$ admits *exact* inference.

The natural strategy is to examine variational methods that improve LBP by processing *regions* (sets of nodes) [41]. In fact we may ignore the variational justification of our algorithm and view it as a *clustered* propagation method, where messages are exchanged amongst regions. As there is no systematic way in the literature to define such regions optimally, we propose a scheme that is well suited to our purposes, by grouping nodes in the shattered network into larger conglomerates, and then running LBP on a set of regions that emerges easily from these conglomerates. In order to group nodes without concerns about directed cycles, we use *factor graphs*; that is, undirected bipartite graphs where circular nodes correspond to variables and rectangular nodes correspond to factors of the joint distribution, and where an edge connects a circular node and a rectangular node when the variable in the former defines the domain of the factor in the latter. Each factor $f_j(\mathbf{X}_j)$ is a conditional probability distribution, where $\mathbf{X}_j$ is a set of indicator functions of grounded relations. Figure 2 (middle) depicts the factor graph for a fragment of Figure 2 (left).

Our proposal is to transform each slice of the grounded network into a small set of factors, and run LBP in the resulting factor graph, while running exact inference *inside* some of the factors. We now elaborate this idea.

Consider the slice for individual $a$. Multiply together every probability distribution in the slice for $a$, except the distributions for indicator variables of restrictions $(\forall r.A)(a)$ and $(\exists r.A)(a)$.[5] The resulting product is called the *factor for $a$*, and denoted by $f_a$. We now construct a factor graph where all factors used to produce $f_a$ are replaced by the factor $f_a$ itself, and run LBP in this compact factor graph.

Denote by $\mathbb{B}_a$ the Bayesian network with all nodes used in $f_a$, and refer to nodes in $\mathbb{B}_a$ that are not grounded at $a$ as *extraneous* node. For instance, if $(\forall r.B)(a)$ belongs to $\mathbb{B}_a$, then $B(b)$ is an extraneous node in $\mathbb{B}_a$. Denote by $\mathbf{E}_a$ the variables in $\mathbb{B}_a$ that correspond to members of the Abox of interest (that is, the evidence in the query).

To simplify the calculation of messages, we introduce an individual $b \notin \mathcal{D}'$. So we have the individuals in $\mathcal{D}'$ plus $b$ and then $x$ standing for all other elements of the domain. Figure 2 (right) depicts a fragment of the resulting factor graph for Example 1, corresponding to the slice of $a_0$. We must propagate the following messages until convergence [41, Eqs. (4), (5)], for a concept $A$ and $a \in \mathcal{D}' \cup \{b\}$:

$$n_{A(a) \to f_{a'}}(A(a)) = \prod_{f \,\in\, G(A(a)) \setminus f_{a'}} m_{f \to A(a)}(A(a)),$$

$$m_{f_{a'} \to A(a)}(A(a)) = \sum_{\mathbf{X}_{a'} \setminus A(a)} f_{a'}(\mathbf{X}_{a'}) \prod_{\hat{A}(\hat{a}) \,\in\, G(f_{a'}) \setminus A(a)} n_{\hat{A}(\hat{a}) \to f_{a'}}(\hat{A}(\hat{a})),$$

where $G(\cdot)$ denotes neighbors in the compact factor graph. Consider the computation of message $m_{f_a \to A(a)}$, sent to node $A(a)$ from $f_a$. The distribution of an extraneous node is either a message $n_{B(a_i) \to f_a}$ from individuals $a_i$, or a message $n_{B(b) \to f_a}$ from the generic individual $b$, or a parameterized message $n_{B(x) \to f_a}$. The only question is how to handle the $N - M$ messages $n_{B(x) \to f_a}$ at once, without actually writing them all. Note that $n_{B(x) \to f_a}$ is actually equal to $n_{B(b) \to f_a}$, because any particular $x$ behaves like $b$. So, we compute the message for $b$ and just use the result for the remaining elements represented by $x$ (this is the reason to keep an individual $b$ separately). As LBP takes these $N - M - 1$ messages related to $x$ to be independent, and we can use this approximation to compute in closed-form the effect of these messages. A simple example: Take $(\forall r.B)(a)$; we need $P(\bigwedge_x r(a,x) \to B(x))$, approximated by $(1 - P(r(a,x))(1 - n_{B(b) \to f_a}))^{N-M}$ (the exponent $N - M$ is needed because the calculation stands for this many elements of the domain). In short, the message $m_{f_a \to A(a)}$ is the result of probabilistic inference in $\mathbb{B}_a$ with a particular set of distributions for extraneous nodes. Similar reasoning yields messages $m_{f_b \to A(b)}$ and $m_{f_a \to A(b)}$. Messages $n_{A(a) \to f_a}$ can be interpreted as $P(A(a), \mathbf{E}_a) \times \prod_{f \,\in\, N(A(a)) \setminus f_a} P(\mathbf{E}_b | A(a))$, where terms are computed in $\mathbb{B}_a$. The point is that messages can be computed as inferences in appropriate Bayesian networks.

As factor $f_a$ and its incoming messages represent a Bayesian network in each iteration, we can use recent algorithms that exploit determinism within probabilistic inference [6, 12, 35]. The use of larger regions leads to improvements in accuracy compared to LBP, while the use of logical inference within regions leads to gains in speed. Of course, it may happen that a particular problem has a t-network so complex that Bayesian network inference fails even for $N = 1$. In this case the method can be applied by breaking factors more finely; that is, by selecting smaller regions for the propagation [41].

### 4.2 Experiments under the Bayesian assumption

We now describe experiments with the proposed inference algorithm. We have used the ACE engine for the probabilistic calculations as it can exploit logical inference [6], running in a dual core Pentium 2GHz with 2GBytes of memory. As we have indicated, the current technology on exact probabilistic inference can only deal with small $N$; it does not seem that an extensive comparison between approximations and exact results is possible at the moment. We analyze a few examples that illustrate well our method.

In the following tables we present results of exact inference, then LBP, then our proposed method. Exact inference always produced results within two seconds but failed for $N > 9$ due to memory exhaustion. The results of LBP (on the grounded network) and parameterized LBP are identical, but their running times are different. We only present results in cases where (grounded) LBP converged in less than 45 minutes. Overall, paremeterized LBP runs in milliseconds and is slightly slower (about 15%) than our proposed method. For the latter we present the result of the inference and the running time in milliseconds (respectively the last two rows of the tables). We always initialize messages with the results of probabilistic inference in a grounded Bayesian network for $N = 1$.

First, we present inferences $P(C(a_0))$ for Example 1. Note first that the proposed method is more accurate than LBP. It is also remarkable that the proposed method *always* converged within *two* iterations, while the number of iterations for LBP varied but was always much larger. We have:

---

[5]In our implementation we decompose each such restriction into conjunction/disjunction of two pieces: one is entirely related to $a$ and its factors are kept within $f_a$; the other is related to other elements of the domain and is kept outside of $f_a$. This increases the cluster related to $f_a$, improving accuracy and performance.

| $N$ | 1 | 5 | 9 | 50 | 200 | 500 |
|---|---|---|---|---|---|---|
| Exact: | 0.5535 | 0.8445 | 0.9210 | — | — | — |
| LBP: | 0.5781 | 0.8558 | 0.9421 | 0.9798 | — | — |
| Proposed: | 0.5335 | 0.8445 | 0.9285 | 0.9739 | 0.4723 | 0.4050 |
| Runtime(ms): | 32 | 30 | 30 | 29 | 29 | 29 |

Our next experiment uses the probabilistic Kangaroo ontology (Section 1), as this ontology leads to very dense grounded networks. We compute $P(\mathsf{Parent}(a_0))$ (again, convergence with LBP took dozens of iterations, while our proposed method always converged almost instantly):

| $N$ | 1 | 5 | 9 | 20 | 50 | 200 |
|---|---|---|---|---|---|---|
| Exact: | 0.1620 | 0.3536 | 0.4481 | — | — | — |
| LBP: | 0.0875 | 0.3196 | 0.4299 | 0.5243 | — | — |
| Proposed: | 0.1620 | 0.3536 | 0.4481 | 0.5268 | 0.5399 | 0.5400 |
| Runtime(ms): | 32 | 32 | 34 | 30 | 36 | 32 |

We finish with a larger terminology containing substantial deterministic information. We have randomly generated a terminology whose t-network has 20 nodes, 15 of which are associated with deterministic definitions and 2 others are associated with restrictions. We compute inference $P(A(a_0))$ where $A(a_0)$ is the node whose inference requires most computation:

| $N$ | 1 | 5 | 9 | 20 | 50 | 200 |
|---|---|---|---|---|---|---|
| Exact: | 0.7240 | 0.7544 | 0.7694 | — | — | — |
| LBP: | 0.6452 | 0.6991 | 0.7257 | 0.7479 | — | — |
| Proposed: | 0.7240 | 0.7544 | 0.7694 | 0.7819 | 0.7840 | 0.7840 |
| Runtime(ms): | 68 | 71 | 69 | 68 | 74 | 70 |

To conclude this section, we note that inferences such as $P(A(a_0)|\mathcal{A}) = \alpha$ are not the only ones our method can produce. To illustrate this, consider the completely different question, "What is the tightest interval $[\underline{\alpha}, \overline{\alpha}]$ such that

$$\forall x \in \mathcal{D} : P(A(x)|\mathcal{A}) \in [\underline{\alpha}, \overline{\alpha}]?"$$

We answer this question by collecting probabilities across the domain $P(A(x)|\mathcal{A})$ after running inference. For instance, consider Example 1 with $N = 9$ and evidence $\mathcal{A} = \{\neg C(a_0), \neg D(a_1), B(a_2), \neg B(a_3)\}$. We then obtain $\forall x \in \mathcal{D} : P(A(x)|\mathcal{A}) \in [0.598, 1]$.

### 4.3 Dropping assumptions

In this very brief section we comment on failures of the Bayesian assumption. Suppose we have the uniqueness assumption but $N$ is countably infinite, thus failing the confined-domain assumption. As already remarked in Section 3, results by Jaeger [23] show that a unique joint distribution exists in this case. To obtain approximate inferences for $N = \infty$, run the propagation method with increasing values of $N$: there will be a point where probabilities raised to $N - M$ become smaller than machine precision, and at that point we reach approximations valid for $N = \infty$. For instance, for the network of Example 1, all messages related to quantifiers are deterministic for $N = 500$, so the approximate value for $P(C(a_0))$ is 0.4050 for $N = \infty$. This is a nice result because in this example one can show that $P(C(a_0))$ is *exactly* 0.4050 for $N = \infty$ [8]. Note that it would be misleading to take the result for $N = 50$ as an approximation for the case $N = \infty$: while the correct value is $P(C(a_0)) = 0.4050$, LBP produces $P(C(a_0)) = 0.9798$ for $N = 50$.

When $N$ is unconstrained we may have different inferences for varying $N$: uniqueness of probabilities is not guaranteed. Thus it is reasonable to analyze unconstrained $N$ together with failure of the uniqueness assumption. We should expect most applications to stay within uniqueness, but there are several reasons why uniqueness may fail in a probabilistic description logics. One reason is that a standard inclusion $C \sqsubseteq D$ does imply $P(C|\neg D) = 0$, but nothing is implied about $P(C|D)$. Another reason is that a precise assessment such as $P(C|A \sqcup B) = \alpha$ does not necessarily constrain probabilities such as $P(C|A \sqcap B)$ down to a single value, so uniqueness may fail. A version of LBP that handles probability intervals can be used when uniqueness fails [8]. This version of LBP, called *L2U* [20] is similar to LBP but it propagates the interval-valued messages derived in the 2U algorithm [15]. We have only two observations to make in this setting. First, as handling unconstrained $N$ is akin to handling probability intervals, messages that are propagated must only be adapted by conducting, for each message, a minimization and a maximization with respect to $N$. Second, we emphasize that our method allows each slice to be processed in isolation, and specific methods for inference with probability intervals can be used inside such factors [9].

To illustrate these techniques, consider again the terminology in Example 1. Suppose we discard the assessment $P(B|A) = \alpha_2$, leaving only the standard inclusion $B \sqsubseteq A$ for $B$. By propagating messages with $N = 10$, we obtain $P(C(a_0)) \in [0.9179, 0.9917]$. If we only impose that $N \leq 20$, leaving $N$ otherwise unconstrained, we obtain $P(C(a_0)) \in [0.2910, 0.9831]$.

## 5 Conclusion

This paper has contributed with novel complexity analysis and algorithms for inference in a family of probabilistic description logics with interpretation-based semantics and assumptions of acyclicity and independence. Variants of Proposition 1 and Theorem 1 should apply to other relational/first-order probabilistic models [16, 34]; we note that currently there are relatively few results on complexity for these models. Also, our analysis produces a meaningful class of **PEXP**-complete problems that may be useful to other applications.

Concerning algorithms, the contributions here center around a variational scheme that, essentially, breaks the shattered network into smaller units, so as to employ logical inference whenever possible. This idea should be useful to other probabilistic logics. Experiments demonstrate that our techniques are better than existing ones and in fact can handle problems of practical significance. The method can be adapted to handle infinite and unconstrained domains, as well as situations where uniqueness is violated. Clearly, many improvements and extensions can be contemplated; in particular, most of this paper focuses on the Bayesian assumption, and more work is to be devoted to situations where this assumption fails.

In this paper we have increased substantially the expressivity of CR$\mathcal{ALC}$, while keeping intact both the worst-case complexity and the basic variational method. We feel justified in starting from CR$\mathcal{ALC}$ as it seems to be a minimal starting point for an interpretation-based probabilistic description logic, as noted in Section 3. A challenge for the future is to include transitivity and cyclic definitions; these features may require a move to undirected t-networks (alas, the usual Markov condition for undirected graphs does not guarantee factorization when deterministic constraints are present [30], so the move to undirected graphs may require substantially new ideas). Other extensions would be desirable, such as unrestricted nominals, but they appear to be quite challenging. As practical applications should be valuable in indicating which constructs are really useful, we plan to devote some effort to applications before we attempt to study all of these extensions.

## Acknowledgements

Thanks to Cassio Polpo de Campos for discussions on computational complexity. This work was partially funded by FAPESP (grants 04/09568-0 and 08/03995-5); the first author is partially supported by CNPq and the second author is supported by FAPESP.

## A  Proof of Theorem 1

Membership: Propositionalize the terminology into an exponentially large Bayesian network and run inference (a **PP**-complete problem in the exponentially large network). Hardness: We resort to bounded domino problems. A *domino system* consists of a finite set $D$ of *tiles* and a pair of *compatibility relations* on $D \times D$, respectively expressing horizontal and vertical constraints between tiles. Some tiles, the *initial conditions*, are assigned to a torus $U(s, t)$, and the torus has a *tiling* if it is possible to cover the whole torus while respecting the constraints and the initial conditions. Börger et al [4, Thm. 6.1.2] show that given a (time/space) bounded Turing machine one can construct a bounded domino system that reproduces its behavior. However their reduction is not *parsimonious* [32, Sec. 18.1] as the number of accepting paths and the number of tilings may differ. A parsimonious reduction can be constructed by enlarging the original bounded Turing machine into a machine that visits every cell in its tape (within the space bound) and that reaches the final accepting state only after a prescribed number of operations (by counting operations via auxiliary counters). Consider then a Turing machine solving some selected **NEXP**-complete problem of size $\mathcal{O}(n)$ in such a way that its translation into a domino system involves a torus of size $2^n \times 2^n$. We wish to encode this domino system using a CR$\mathcal{ALC}$ terminology. Tobies shows how to encode such a torus $U(2^n, 2^n)$ with $\mathcal{ALCQ}$ plus several cardinality constraints [39]. We can adapt his construction to $\mathcal{ALC}$ with a single cardinality constraint $N = 2^n \times 2^n$. To do so, we take the definitions of concepts $C_{0,0}$, $D_{\text{east}}$ and $D_{\text{north}}$ exactly as in Tobies' work; additionally, the constraints $(\exists C_{0,0})$, $(\forall D_{\text{east}})$, $(\forall D_{\text{north}})$, $(\forall (\exists \text{east}.\top))$, $(\forall (\exists \text{north}.\top))$, where: $\top$ denotes $A \sqcup \neg A$ for some $A$ not in the terminology; $(\exists C)$ denotes $\exists x \in \mathcal{D} : C(x)$; $(\forall C)$ denotes $\forall x \in \mathcal{D} : C(x)$. Using these constraints it is possible to construct an isomorphism between any model of the terminology and a torus $U(2^n, 2^n)$ (by Tobies' finite induction argument plus the fact that the constraint on $N$ forces a single element to be associated with each point in the torus [39, pp. 205-206]).

The key insight now is to "simulate" constraints such as $(\forall C)$ and $(\exists C)$ using probabilities. To do so, assign probability $1/2$ to every free concept/role in Tobies's construction ($P(X_i) = P(Y_i) = P(\text{east}) = P(\text{north}) = 1/2$) and introduce a new role $r$ and assessment $P(r) = 1$. The probability $P(C'(a_0))$ for a concept $C' \equiv \forall r.C \sqcap \exists r.C_{0,0}$, where $C$ is the conjunction $D_{\text{east}} \sqcap D_{\text{north}} \sqcap (\exists \text{east}.\top) \sqcap (\exists \text{north}.\top)$, is the probability that a torus is built under the probabilistic assessments. We must still encode the compatibility relations and the initial conditions on the torus; this is done exactly as in Tobies' construction, by using concepts $C_d$, $C_{i,0}$ and associated definitions and constraints of the form $(\forall C)$. We again simulate these latter constraints probabilistically: introduce $P(C_d) = 1/2$ for all $C_d$, define $\hat{C} \equiv C' \sqcap \forall r.C''$ where $C''$ is the conjunction of all concepts used in the compatibility constraints and initial conditions. Now $\gamma \doteq P(\hat{C}(a_0))$ is the probability that a torus satisfying all conditions is built. If we can recover the number of tilings of the torus from $\gamma$, we obtain the number of accepting computations of the original exponentially-bounded Turing machine. Note that $\gamma \times 2^\delta$ is the number of truth assignments that build the torus satisfying horizontal and vertical relations and initial conditions, where $\delta$ is the number of logically independent random variables in the grounding of the terminology (we have $\delta = 2^{2N}(2N + 2^{2N+1} + |D|)$). This number is *not* equal to the number of tilings of the torus; to produce the number of tilings of the torus, we must compute $\gamma \times 2^\delta / 2^{2n}!$, where we divide the number of satisfying truth assignments by the number of repeated tilings. Consequently we obtain the number of accepting computations of the original Turing machine just by processing the inference $P(\hat{C}(a_0))$. This shows that $\mathsf{INF}_B(Q)$ is **PEXP**-hard.□

## References

[1] F. Baader, D. Calvanese, D.L. McGuinness, D. Nardi, and P.F. Patel-Schneider. *Description Logic Handbook*. Cambridge University Press, 2002.

[2] F. Baader. Terminological cycles in a description logic with existential restrictions. *Int. Joint Conf. on AI*, pages 325–330, 2003.

[3] F. Bacchus. *Representing and Reasoning with Probabilistic Knowledge: A Logical Approach*. MIT Press, 1990.

[4] E. Börger, E. Grädel, and Y. Gurevich. *The Classical Decision Problem*. Springer, 1997.

[5] H. Buhrman, L. Fortnow, and T. Thierauf. Nonrelativizing separations. *Proc. of IEEE Complexity*, pages 8–12, 1998.

[6] M. Chavira and A. Darwiche. On probabilistic inference by weighted model counting. *Artificial Intelligence*, 172(6-7):772–799, 2008.

[7] P. C. G. Costa and K. B. Laskey. PR-OWL: A framework for probabilistic ontologies. *Conf. on Formal Ontology in Information Systems*, 2006.

[8] F. G. Cozman and R. B. Polastro. Loopy propagation in a probabilistic description logic. *Int. Conf. on Scalable Uncertainty Management* (LNAI 5291), pages 120–133. Springer, 2008.

[9] C. Polpo de Campos and F. G. Cozman. The inferential complexity of Bayesian and credal networks. *Int. Joint Conf. on AI*, pages 1313–1318, Edinburgh, United Kingdom, 2005.

[10] C. D'Amato, N. Fanizzi, and T. Lukasiewicz. Tractable reasoning with Bayesian description logics. *Int. Conf. on Scalable Uncertainty Management* (LNAI 5291), pages 146–159, 2008.

[11] R. de Salvo Braz, E. Amir, and D. Roth. Lifted first-order probabilistic inference. *Int. Joint Conf. on AI*, 2006.

[12] R. Dechter and R. Mateescu. AND/OR search spaces for graphical models. *Artificial Intelligence*, 171:73–106, 2007.

[13] Z. Ding, Y. Peng, and R. Pan. BayesOWL: Uncertainty modeling in semantic web ontologies. *Soft Computing in Ontologies and Semantic Web*, pages 3–29. Springer, Berlin/Heidelberg, 2006.

[14] M. Dürig and T. Studer. Probabilistic ABox reasoning: preliminary results. *Description Logics*, pages 104–111, 2005.

[15] E. Fagiuoli and M. Zaffalon. 2U: An exact interval propagation algorithm for polytrees with binary variables. *Artificial Intelligence*, 106(1):77–107, 1998.

[16] L. Getoor and B. Taskar. *Introduction to Statistical Relational Learning*. MIT Press, 2007.

[17] J. Y. Halpern. *Reasoning about Uncertainty*. MIT Press, Cambridge, Massachusetts, 2003.

[18] J. Heinsohn. Probabilistic description logics. *Conf. Uncertainty in AI*, page 311-318, 1994.

[19] I. Horrocks, P. F. Patel-Schneider, F. van Harmelen. From SHIQ and RDF to OWL: The making of a web ontology language. *Journal of Web Semantics*, 1(1):7–26, 2003.

[20] J. S. Ide and F. G. Cozman. Approximate algorithms for credal networks with binary variables. *Int. Journal of Approximate Reasoning*, 48(1):275–296, 2008.

[21] M. Jaeger. Probabilistic reasoning in terminological logics. *Principles of Knowledge Representation*, pages 461–472, 1994.

[22] M. Jaeger. Relational Bayesian networks. *Conf. Uncertainty in AI*, pages 266–273, San Francisco, California, 1997.

[23] M. Jaeger. Reasoning about infinite random structures with relational Bayesian networks. *Knowledge Representation*, San Francisco, California, 1998. Morgan Kaufman.

[24] M. Jaeger. On the complexity of inference about probabilistic relational models. *Artificial Intelligence*, 117(2):297–308, 2000.

[25] D. Koller, A. Y. Levy, and A. Pfeffer. P-CLASSIC: A tractable probablistic description logic. *AAAI Conf. on AI*, pages 390–397, 1997.

[26] H. E. Kyburg Jr. and C. M. Teng. *Uncertain Inference*. Cambridge University Press, 2001.

[27] M. L. Littman, S. M. Majercik, and T. Pitassi. Stochastic Boolean satisfiability. *Journal of Automated Reasoning*, 27(3):251–296, 2001.

[28] T. Lukasiewicz. Expressive probabilistic description logics. *Artificial Intelligence*, 172(6-7):852–883, April 2008.

[29] T. Lukasiewicz and U. Straccia. Managing uncertainty and vagueness in description logics for the semantic web. *Journal of Web Semantics*, 6(4):291–308, November 2008.

[30] J. Moussouris. Gibbs and Markov random systems with constraints. *Journal of Statistical Physics*, 10(1):11–33, 1974.

[31] C. H. Papadimitriou. Games against nature (extended abstract). *24th Annual Symposium on Foundations of Computer Science*, pages 446–450, 1983.

[32] C. H. Papadimitriou. *Computational Complexity*. Addison-Wesley Publishing, 1994.

[33] R. B. Polastro and F. G. Cozman. Inference in probabilistic ontologies with attributive concept descriptions and nominals. *4th Int. Workshop on Uncertainty Reasoning for the Semantic Web*, Karlsruhe, Germany, 2008.

[34] L. De Raedt. *Logical and Relational Learning*. Springer, 2008.

[35] T. Sang, P. Beame, and H. Kautz. Solving Bayesian networks by weighted model counting. *AAAI Conf. on AI*, Pittsburgh, PA, 2005.

[36] M. Schmidt-Schauss and G. Smolka. Attributive concept descriptions with complements. *Artificial Intelligence*, 48:1–26, 1991.

[37] F. Sebastiani. A probabilistic terminological logic for modelling information retrieval. *17th Conf. on Research and Development in Information Retrieval*, pages 122–130, Dublin, Ireland, 1994. Springer-Verlag.

[38] P. Singla and P. Domingos. Lifted first-order belief propagation. *AAAI Conf. on AI*, 2008.

[39] S. Tobies. The complexity of reasoning with cardinality restrictions and nominals in expressive description logics. *Journal of Artificial Intelligence Research*, 12:199–217, 2000.

[40] H. Veith. Languages represented by Boolean formulas. *Information Processing Letters*, pages 251–256, 1997.

[41] J. S. Yedidia, W. T. Freeman, and Y. Weiss. Constructing free energy approximations and generalized belief propagation algorithms. *IEEE Transactions on Information Theory*, 51:2282–2312, 2005.